\documentclass[sigconf]{acmart}
\usepackage{graphicx}
\usepackage{subcaption}
\usepackage{CJKutf8}
\usepackage{algorithm}
\usepackage{algorithmic}
\usepackage{listings}
\usepackage{xcolor}

\definecolor{codegreen}{rgb}{0,0.6,0}
\definecolor{codegray}{rgb}{0.5,0.5,0.5}
\definecolor{codepurple}{rgb}{0.58,0,0.82}
\definecolor{backcolour}{rgb}{0.95,0.95,0.92}

\lstdefinestyle{mystyle}{
    commentstyle=\color{codegreen},
    keywordstyle=\color{magenta},
    numberstyle=\tiny\color{codegray},
    stringstyle=\color{codepurple},
    basicstyle=\ttfamily\footnotesize,
    breakatwhitespace=false,         
    breaklines=true,                 
    captionpos=b,                    
    keepspaces=true,                 
    numbers=left,                    
    numbersep=5pt,                  
    showspaces=false,                
    showstringspaces=false,
    showtabs=false,                  
    tabsize=2
}

\AtBeginDocument{%
  \providecommand\BibTeX{{%
    \normalfont B\kern-0.5em{\scshape i\kern-0.25em b}\kern-0.8em\TeX}}}

\setcopyright{acmcopyright}
\copyrightyear{2022}
\acmYear{2022}
\acmDOI{10.1145/1122445.1122456}

\acmConference[Arxiv]{Arxiv}{2022}{Arxiv}

\def\m{{\bf m}}

\def\p{{\bf p}}

\def\q{{\bf q}}
\def\x{{\bf x}}

\def\z{{\bf z}}
\def\0{{\bf 0}}
\def\1{{\bf 1}}

\def\muu{\mbox{\boldmath$\mu$\unboldmath}}

\def\nuu{\boldsymbol\nu}

\def\etal{{\em et al.\/}\,}

\begin{document}

\title{MACE: An Efficient Model-Agnostic Framework for Counterfactual Explanation}

\author{Wenzhuo Yang, Jia Li, Caiming Xiong, Steven C.H. Hoi}
\affiliation{%
  \institution{Salesforce Research}
  \country{}}
\email{{wenzhuo.yang, jia.li, cxiong, shoi}@salesforce.com}

\if 0 
\author{Wenzhuo Yang}
\affiliation{%
  \institution{Salesforce Research}
  \country{Singapore}}
\email{wenzhuo.yang@salesforce.com}

\author{Jia Li}
\affiliation{%
  \institution{Salesforce Research}
  \city{Palo Alto}
  \state{CA}
  \country{USA}}
\email{jia.li@salesforce.com}

\author{Caiming Xiong}
\affiliation{%
  \institution{Salesforce Research}
  \city{Palo Alto}
  \state{CA}
  \country{USA}}
\email{cxiong@salesforce.com}

\author{Steven C.H. Hoi}
\affiliation{%
  \institution{Salesforce Research}
  \country{Singapore}}
\email{shoi@salesforce.com}
\fi


\begin{abstract}
Counterfactual explanation is an important Explainable AI technique to explain machine learning predictions. Despite being studied actively, existing optimization-based methods often assume that the underlying machine-learning model is differentiable and treat categorical attributes as continuous ones, which restricts their real-world applications when categorical attributes have many different values or the model is non-differentiable. To make counterfactual explanation suitable for real-world applications, we propose a novel framework of \textbf{M}odel-\textbf{A}gnostic \textbf{C}ounterfactual \textbf{E}xplanation (MACE), which adopts a newly designed pipeline that can efficiently handle non-differentiable machine-learning models on a large number of feature values. in our MACE approach, we propose a novel RL-based method for finding good counterfactual examples and a gradientless descent method for improving proximity. Experiments on public datasets validate the effectiveness with better validity, sparsity and proximity. For practical adoption, we developed an easy-to-use counterfactual explanation toolkit based on MACE for our internal use cases, and a demo video can be found here\footnote{https://youtu.be/X0FX-jrsJx4}.
\end{abstract}


\maketitle

\section{Introduction}

%Machine learning (ML) models have been widely adopted in a wide range of real-world AI applications, such as computer vision~\cite{Krizhevsky2012}, natural language processing~\cite{Chang2019}, data mining and recommender systems~\cite{Cheng2016}. 
%As the adoption of these models grows rapidly, 
With the rapidly growing adoptions of machine-learning (ML) models in real-world applications, e.g., \cite{Krizhevsky2012,Chang2019,Cheng2016}, their algorithmic decisions can potentially have a huge societal impact, especially for some application domains such as healthcare, education, and finance. Many ML models especially deep learning based models work as black-box models that lack explainability and thus inhibit their adoptions in some critical applications and hamper the trust in machine learning/AI systems. To address these challenges, Explainable AI (XAI) is an emerging field in machine learning and AI, aiming to explain how those black-box decisions of ML systems are made. Such explanations can improve the transparency, persuasiveness, and trustworthiness of AI systems and help AI developers to debug and improve model performance, e.g.,~\cite{doshivelez2017rigorous,Kusner2017,Lipton2016,Lundberg2020}. 
 
\begin{figure}[!ht]
\center
\includegraphics[width=1.0\linewidth]{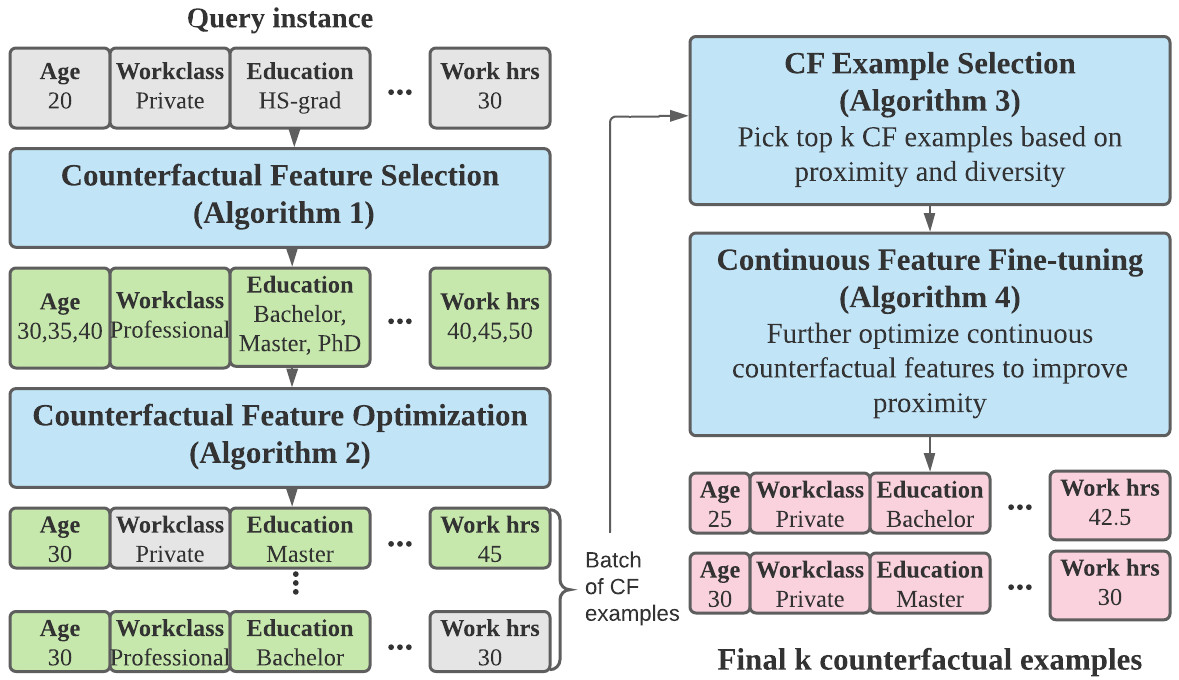}
%\vspace{-0.1in}
\caption{Overview of the proposed MACE framework.}
\label{fig:overview}%\vspace{-0.1in}
\end{figure}

Among various explanation techniques, counterfactual explanation (CE) is a promising tool that can not only explain the outcome of a model's decision but also provide insight on how to change the outcome in the future. Formally, CE interprets a model's decision on a query instance by generating counterfactual examples that have minimal changes w.r.t the query instance’s features to yield a predefined output~\cite{Wachter2017CounterfactualEW,Moraffah2020,Byrne2019}. It describes a causal situation in the form ``Y would not have occurred if X had not occurred''. For example, consider a person who submitted a credit card application and was rejected by a financial AI system. The applicant may require to explain the decision. CE can explain why the application was rejected and tell what he could have done to change this decision, e.g., ``if your income was higher than \$50k/year, your application would have been approved''. Unlike the explanations based on feature importance~\cite{Ribeiro2016,Lundberg2017}, CE allows users to explore “what-if” scenarios and obtain more actional insights of the underlying ML models.

Recently, many approaches have been proposed for counterfactual explanations, e.g.,~\cite{Wachter2017CounterfactualEW,Mothilal2020,pmlr-v97-goyal19a,ates2020counterfactual,Dhurandhar2018,Russell2019,Sharma2020,mahajan2019preserving,Hendricks2018,Looveren2019,Kanamori2020,vanderwaa2018contrastive,dhurandhar2019model,karimi2020modelagnostic}. They are often optimization-based and most are solved based on continuous optimization, e.g., \cite{Wachter2017CounterfactualEW,Rory2018,Mothilal2020,Dhurandhar2018} assume that the underlying ML model is differentiable and static, i.e., the model is fixed and easy to compute the gradients w.r.t. its inputs. Such assumption is valid for neural networks but invalid for tree boosting models such as XGBoost~\cite{Chen2016}. Besides, these methods do not handle categorical features with many different values well~\cite{molnar2019}. For example, the DiCE method in \cite{Mothilal2020} converts categorical features into one-hot encoding representations and then treats them as continuous features. This relaxation introduces many new variables, which increases computational cost a lot and downgrades the sparsity of the counterfactual examples. To ensure real-world plausibility of the counterfactual examples, additional constraints are imposed on features especially for continuous features, making the original optimization problem harder to solve. These drawbacks make those approaches hard to apply in real-world applications.

To address these issues, we propose a novel and efficient Model-Agnostic Counterfactual Explanations (MACE) framework. As a summary, our work makes the following key contributions:
\begin{enumerate}
\item We propose a novel framework for generating counterfactual examples (as shown in Figure \ref{fig:overview}) via four stages: counterfactual feature selection, counterfactual feature optimization, counterfactual example selection, and continuous feature fine-tuning. It achieves better sparsity, proximity and running time per query than the existing approaches. 
%Our pipeline includes four stages : counterfactual feature selection, counterfactual feature optimization, counterfactual example selection, and continuous feature fine-tuning.
\item For counterfactual feature optimization, we propose a method based on REINFORCE \cite{Williams92simplestatistical,Sutton2000} for generating good counterfactual examples, and demonstrate its connections to the previous approaches. For continuous feature fine-tuning, we propose an efficient gradientless descent method to further improve the proximity of the final counterfactual examples.
\item The experiments show the effectiveness of our method and the ability to handle time series classification tasks. Our approach runs much faster than DiCE \cite{Mothilal2020} and CERTIFAI \cite{Sharma2020}, making it practical for adoption in real-world applications.
\item For practical adoption, we developed an easy-to-use counterfactual explanation toolkit to support our internal use cases. More details about our toolkit can be found in Appendix.
\end{enumerate}

\section{Related Work}
In literature, a variety of approaches have been proposed to generate counterfactual explanations, e.g.,~\cite{Wachter2017CounterfactualEW,Mothilal2020,pmlr-v97-goyal19a,ates2020counterfactual,Dhurandhar2018,Russell2019,Sharma2020,mahajan2019preserving,Hendricks2018,Looveren2019,Kanamori2020,vanderwaa2018contrastive,dhurandhar2019model,karimi2020modelagnostic}. Most of them were solved based on relaxed continuous optimization approaches. Wachter \etal~\cite{Wachter2017CounterfactualEW} proposed an optimization formulation by minimizing the distance between the query instance and its counterfactual examples in the feature space, and the mean squared error between the model prediction of those counterfactual examples and the desired label. Based on this promising optimization approach, Grath \etal \cite{Rory2018} developed two weighting strategies to generate more interpretable counterfactuals in a high dimensional setting, e.g., credit application prediction via off-the-shelf black-box classifiers. Mothilal \etal \cite{Mothilal2020} proposed DiCE as a new formulation for generating a diverse set of counterfactual explanations based on determinantal point processes and provided metrics that enable comparison of counterfactual-based methods to evaluate the actionability of counterfactuals. Sharma \etal \cite{Sharma2020} proposed CERTIFAI for CE optimized by a genetic algorithm which is sub-optimal and slow to scale in practice.  

Besides dealing with tabular data, counterfactual explanation can also be applied for other types of data. Liu \etal~\cite{Shusen2019} proposed a generative and actionable
counterfactual explanations generation framework based on the formulation proposed by Wachter \etal~\cite{Wachter2017CounterfactualEW} for image classification tasks. Goyal \etal~\cite{pmlr-v97-goyal19a} developed a technique to generate counterfactual visual explanations by detecting spatial regions in a query image and a distractor image such that swapping those regions results in predicting the query image as the class of the distractor image. This technique is an optimization approach by solving the minimum-edit counterfactual problem. Dhurandhar \etal~\cite{Dhurandhar2018} proposed another method that provides counterfactual explanations for image classification tasks by finding what should be minimally and sufficiently present and what should be minimally and necessarily absent to justify its classification. Hendricks \etal~\cite{Hendricks2018} proposed a CE approach for image classification, which takes an image and counterfactual class as the inputs and outputs text sentences to explain why the image does not belong to the counterfactual class. Ates \etal~\cite{ates2020counterfactual} developed a greedy CE approach for supervised learning tasks with multivariate time series data and applied it to HPC system analytics. 

\section{MACE: Counterfactual Generation}

\subsection{Overview}
The inputs of a CE task include a trained machine learning model $f(\cdot)$ and a query instance $\x \in \mathbb{R}^d$ with predicted label $y$ to explain. The goal is to generate a counterfactual example $\x' \in \mathbb{R}^d$ such that its predicted label $y'$ is different from $y$. Most existing approaches, e.g., \cite{Wachter2017CounterfactualEW,Mothilal2020}, formulate this problem as an optimization problem by minimizing the distance between $\x$ and $\x'$ subject to that $f(\x) \neq y$. This problem is usually hard to solve due to the complexity of the machine learning model $f(\cdot)$, the existence of categorical features in $\x$ and the real-world feasibility constraints on $\x'$. The common ideas leveraged by those approaches to address these issues are as follows: 1) one assumes that the model is differentiable, 2) the categorical features are treated as continuous ones by converting them into one-hot encoding representations, and 3) additional constraints are imposed to ensure that $\x'$ is real-world feasible.

The assumptions and relaxations mentioned above lead to promising results but there still exists a big gap for practical applications. Widely applied tree boosting models such as XGBoost don't have closed-form solutions for gradients, making those approaches difficult to apply on tree boosting models. For a categorical feature, when the number of its possible feature values is large, this relaxation makes the optimization problem much harder due to introducing much more decision variables and constraints, leading to high latency per query. To handle black-box models and generate counterfactual examples efficiently, we propose a novel model-agnostic framework for counterfactual explanation (MACE). For ease of presentation, we consider counterfactual explanation for binary-class classification models while our approach can be easily extended to deal with multiple-class classification models. 

Figure \ref{fig:overview} presents the system overview of our MACE approach. Given a query instance, counterfactual feature selection picks a subset of feature columns and values that can potentially change the prediction. Counterfactual feature optimization determines the best way to generate a batch of candidate CF examples based on the selected features. These candidates are then sent to the CF example selector which ranks them based on proximity and diversity. Finally, the continuous counterfactual features are fine-tuned to make the examples more actionable, e.g., ``work hrs = 45'' to ``work hrs = 42.5''.

\subsection{Counterfactual Feature Selection}

The implicit assumption for the optimization approaches such as \cite{Wachter2017CounterfactualEW,Mothilal2020} is that the domain of a variable related to a categorical feature in $\x'$ contains all possible feature values and the domain of a variable related to a continuous feature is either $\mathbb{R}$ or defined by a real-world feasibility constraint. These two domains are usually too broad, complicating the optimization procedure for finding $\x'$.

For example, let's consider a binary-class classification task of predicting if the income of a person exceeds \$50K/yr based on the attributes such as age, education, working hours. The classifier can be trained with the census income data. From basic statistical analysis, one knows that if his/her age is above 30 or his/her education is higher than a bachelor's degree or he/she has longer working hours, he/she probably has an income exceeding \$50K/yr. Given an instance with ``age = 20, education = high school, working hours = 30, predicted income < \$50K/yr'' to explain, instead of considering the whole domains for age, education and working hours when generating counterfactual examples, it is better to narrow the domains down to ``age $\geq$ 30, education $\in$ [bachelor, master, doctorate], working hours $\geq$ 40''. The new domains simplify the optimization and can generate more reliable explanations since they reduce the search space and guarantee practical feasibility. For counterfactual feature selection, we only care about a rough estimate for continuous features in $\x'$, e.g., ``working hours = range [45,50]''. Thus, instead of converting categorical features into continuous ones by continuous optimization approaches, we convert continuous features into categorical ones by discretizing them into $k$ bins.

Algorithm \ref{alg_nn_search} shows the kNN search approach for counterfactual feature selection. The data for training the classifier is divided into two groups according to the predicted labels. For each group, a search index is built for fast kNN search, where the distance metric applied here is the euclidean distance. We apply one-hot encoding to categorical features while applying ordinal encoding to continuous features for preserving the information of the closeness, e.g., age of 20-25 is closer to 25-30 than 45-50. The approximate kNN search algorithms such as HNSW \cite{Malkov2020} can be applied here, which can handle millions of data points efficiently. Training data $\mathcal{D}$ in Algorithm \ref{alg_nn_search} can be either a subset of the examples or all the examples for training the classifier.
\begin{algorithm}
\caption{NN search for counterfactual feature selection}
\label{alg_nn_search}
\begin{algorithmic}
    \REQUIRE \ Training data $\mathcal{D}$, query instance $\x$ with predicted label $y$, number of nearest neighbors $K$, number of selected feature columns $s$, number of selected values for each feature $m$.
    \ENSURE ~~
        \STATE \textbf{Initialize the kNN search indices for all queries}: 
        \STATE $1)$ For each label $l \in \{0,1\}$, build kNN search index $tree_l$ with data $\{\x_i | y_i = l \text{ for } (\x_i, y_i) \in \mathcal{D}\}$;
        \STATE \textbf{For a single query instance}: 
        \STATE $2)$ Let $y' = 1 - y$ and find the $K$ nearest neighbors via $tree_{y'}$. Denote them by $\mathcal{Z} = \{\z_1,\cdots,\z_K\}$;
        \STATE $3)$ Initialize $col\_count$ that counts the number of different feature columns in $\mathcal{Z}$, e.g., $col\_count[c] = |\{i : \z_i[c] \neq \x[c] \text{ for } i = 1, \cdots, K\}|$;
        \STATE $4)$ Initialize $val\_count$ that counts the number of different feature values in $\mathcal{Z}$, e.g., $val\_count[c][v]$ is the number of appearances of value $v$ for feature column $c$;
        \FOR {$c$ in features}
            \STATE $5)$ $col\_count[c] \mathrel{+}= 1$ if $\x_c \neq \z_{i,c}$ for each $\z_i \in \mathcal{Z}$;
            \STATE $6)$ $val\_count[c][\z_{i,c}] \mathrel{+}= 1$ if $\x_c \neq \z_{i,c}$ for each $\z_i \in \mathcal{Z}$;
        \ENDFOR
        \STATE $7)$ Sort the feature columns in descending order w.r.t. the count $col\_count[c]$ and pick the top $s$ feature columns. Denote the selected feature columns by $\mathcal{C}$;
        \STATE $8)$ For each $c$ in $\mathcal{C}$, sort the elements in $val\_count[c]$ in descending order w.r.t. $val\_count[c][v]$ and pick the top $m$ feature values. Denote the selected feature values by $\mathcal{V}(c)$.
        \STATE $9)$ Return $\mathcal{C}$ and $\mathcal{V}(\mathcal{C})=\{\mathcal{V}(c)|c \in \mathcal{C}\}$.
\end{algorithmic}
\end{algorithm}

Given the $K$ nearest neighbors of $\x$ in the opposite class $y' = 1 - y$, the frequencies of different feature columns and feature values from $\x$ are computed. We choose the top $s$ feature columns and the top $m$ values for each feature column. This algorithm generates at most $s \times m$ candidate features and runs efficiently. Algorithm \ref{alg_nn_search} allows users to specify actionable feature columns or the columns they care about most. Before constructing the search index, instead of taking all the features in $\x$ into account, it only needs to consider a subset of the features specified by the users and builds a search index based on it.

\subsection{Counterfactual Feature Optimization}

The second step is to generate a batch of good counterfactual examples given the candidate features $\mathcal{C}$ and $\mathcal{V}(\mathcal{C})$ generated by counterfactual feature selection. 

Suppose that $\mathcal{C} = \{c_1,\cdots,c_s\}$ and $\mathcal{V}(c_i) = \{v_1,\cdots,v_m\}$ for $i=1,\cdots,s$. This problem can be formulated as a reinforcement learning problem. The environment is defined by classifier $f(\cdot)$, instance $(\x,y)$, and candidate features $\mathcal{C}$ and $\mathcal{V}(\mathcal{C})$. The action is defined by several feature column and value pairs to modify to construct $\x'$. The state is $\x'$ by taking an action on $\x$. The reward function is the prediction score of label $y' = 1- y$. For each round $t$, an action $a_t$ is taken to construct $\x_t'$ and then the environment sends the feedback reward $r_t = f_{1-y}(\x_t')$. Let $\pi_{\theta}(\cdot)$ be a stochastic policy parameterized by $\theta$ for choosing actions. The objective is to find the optimal policy that maximizes the cumulative reward:
\begin{equation}\label{eq_1}
\begin{small}
    \max_{\theta}J(\theta) := \mathbb{E}_{\pi_{\theta}}\left[\sum_{t=1}^T r_t\right],\ s.t. \text{ \# of select features by } \pi_{\theta} \leq w.
\end{small}
\end{equation}
Note that a sparsity constraint is imposed on policy $\pi_{\theta}$, which ensures that the number of the selected feature columns to modify cannot be larger than constant $w$.

We now define the formulation for $\pi_{\theta}$. Let $\muu=(\mu_1,\cdots,\mu_s) \in \{0,1\}^s$ be a random vector whose $i$th element indicates whether the $i$th feature column is selected or not, e.g., $\mu_i=1$ means it is selected, and $\mathcal{P}$ be the probability distribution of $\muu$. For each $c \in \mathcal{C}$, let $\nu_c \in \{1,\cdots,m\}$ be a random variable indicating which feature value is selected for the $c$th feature, $\nuu$ be the random vector $(\nu_1,\cdots,\nu_c)$ and $\mathcal{Q}_c$ be the probability distribution of $\nu_c$.

With the notation above, an action can be represented by $\muu$ and $\nuu$, e.g., $a = \{(\mu_1=1,\nu_1=2), (\mu_2=0,\nu_2=None), (\mu_3=1,\nu_3=1)\}$, meaning that the first and third feature columns are set to values 2 and 1, respectively. Then policy $\pi_{\theta}$ can be defined by
\begin{equation}\label{eq_2}
    \pi_{\theta}(a) = \mathcal{P}(\muu)\prod_{c=1}^s\mathcal{Q}_c(\nu_c)^{\mathbf{1}[\mu_c = 1]},
\end{equation}
where $\mathbf{1}[\cdot]$ is the indicator function which equals 1 if the condition is met or 0 otherwise. We assume that $\mathcal{P}$ is constructed by $s$ independent Bernoulli distribution parameterized by $p_c$ and $\mathcal{Q}_c$ is defined by the softmax function parameterized by $\q_c \in \mathbb{R}^m$, namely,
\begin{equation}\label{eq_3}
\mathcal{P}(\muu) = \prod_{c=1}^s\text{Bernoulli}(p_c, \mu_c),\ 
\mathcal{Q}_c(\nu) = \frac{e^{q_{c,\nu}}}{\sum_{i=1}^m e^{q_{c,i}}},
\end{equation}
where $q_{c,i}$ is the $i$th element of $\q_c$, then $\theta = (p_1,\cdots,p_s,\q_1,\cdots,\q_s)$. Under this assumption, the sparsity constraint in \eqref{eq_1} is equivalent to $\|\p\|_0 \leq w$ where $\p = (p_1,\cdots,p_s)$ and $\|\cdot\|_0$ is the number of nonzero elements. Therefore, the optimization problem \eqref{eq_1} can be reformulated as
\begin{equation}\label{eq_4}
    \max_{\theta}\mathbb{E}_{\pi_{\theta}}\left[\sum_{t=1}^T r_t\right],\ s.t.\ \|\p\|_0 \leq w,
\end{equation}
where $\pi_{\theta}$ is defined by \eqref{eq_2}.
The problem with the $l_0$-norm constraint is hard to solve. Therefore, we do the following relaxation:
\begin{equation}\label{eq_5}
    \min_{\theta}-\mathbb{E}_{\pi_{\theta}}\left[\sum_{t=1}^T r_t\right] + \lambda_1\|\p\|_1 + \lambda_2 \sum_{c=1}^s p_c\log p_c,
\end{equation}
where $\lambda_1$ and $\lambda_2$ are constant.
The first regularization term $\|\p\|_1$ is used to encourage sparse solutions, and the second regularization term is the probability entropy that encourages exploration during minimization via policy gradient. To solve this problem, we apply the REINFORCE algorithm \cite{Williams92simplestatistical,Sutton2000}.

Algorithm \ref{alg_feature_optimization} provides the details about the RL-based counterfactual feature optimization and counterfactual example generation. It obtains the optimal policy $\pi_{\theta^*}$ by solving the optimization problem \eqref{eq_5}. Based on $\pi_{\theta^*}$, we first apply a greedy method to construct the baseline counterfactual example $\x_0'$, i.e., we pick the top $w$ feature columns with the highest values of $\p^*$ and the top feature value $v_c^*$ with the highest value of $\q_c$ for each column $c$, and then the features in $\x$ are modified iteratively until the predicted label becomes $y'$. We then samples a batch of counterfactual examples $\{\x_1,\cdots,\x_B'\}$ based on policy $\pi_{\theta^*}$ and combine them with $\x_0'$. Note that $\mathcal{F}^*$, which provides the best counterfactual features based on $\pi_{\theta^*}$, can also be utilized for explaining predictions as demonstrated in our experiments.

\begin{algorithm}
\caption{The RL-based counterfactual feature optimization}
\label{alg_feature_optimization}
\begin{algorithmic}
    \REQUIRE \ Classifier $f(\cdot)$, query instance $(\x,y)$, the selected features $\mathcal{C}$ and $\mathcal{V}(\mathcal{C})$, sparsity parameter $w$ and batch size $B$.
    \ENSURE ~~
        \STATE $1)$ Solve Problem \eqref{eq_5} by the REINFORCE algorithm to obtain the optimal solution $\theta^*=\{p_1^*,\cdots,p_s^*,\q_1^*,\cdots,\q_s^*\}$;
        \STATE $2)$ Sort $(p_1^*,\cdots,p_s^*)$ in descending order and pick the top $w$ ones with the highest values. Denote the selected features by $\mathcal{C}^*$;
        \STATE $3)$ For each $c$ in $\mathcal{C}^*$, pick the feature value $v_c^*$ such that $v_c^* = \arg\max_i q_{c,i}^*$. Set $\mathcal{F}^* = \{(c,v_c^*)\ |\ c \in \mathcal{C}^*\}$ and $\x_0'=\x$;
        \STATE $4)$ Sort $\mathcal{F}^*$ in descending order w.r.t. $p_c^*$ for $(c,\_)$ in $\mathcal{F}^*$;
        \STATE $5)$ For $(c,v)$ in the sorted $\mathcal{F}^*$: set $\x_0'[c] = v$, and break the loop if $f_{1-y}(\x_0') > 0.5$;
        \STATE $6)$ Sample $B$ actions based on policy $\pi_{\theta^*}$ and construct the corresponding examples $\{\x_1'\cdots,\x_B'\}$;
        \STATE $7)$ Return $\mathcal{E} = \{\x_0'\}\cup\{\x_i' : f_{1-y}(\x_i') > 0.5\text{ for } i=1,\cdots,B\}$;
\end{algorithmic}
\end{algorithm}

Our RL-based method has a strong connection with the previous optimization approaches \cite{Wachter2017CounterfactualEW}. Suppose that $\pi_{\theta}$ is a deterministic policy defined by Equation \eqref{eq_2}. Now instead of defining $\mathcal{P}$ and $\mathcal{Q}_c$ as the probability distributions shown in \eqref{eq_3}, we here define $\mathcal{P}$ and $\mathcal{Q}_c$ as follows:
\begin{equation}\label{eq_6}
    \mathcal{P}(\mu) = \prod_{c=1}^s \mathbf{1}[p_c = \mu_c],\ 
    \mathcal{Q}_c(\nu) = \mathbf{1}[q_{c,\nu} = 1],
\end{equation}
where $\mathbf{1}[\cdot]$ is the indicator function, $p_c \in \{0,1\}$ and $\q_c \in \mathbb{R}^m$ is a one-hot vector for $c=1,\cdots,s$. Clearly, the definition of $\mathcal{P}$ and $\mathcal{Q}_c$ here is an unrelaxed formulation of \eqref{eq_3}. The following theorem shows the connection with the previous approaches:
\begin{theorem}\label{th_1}
Given the instance $(\x,y)$ to explain and the domains defined by $\mathcal{C}$ and $\mathcal{V}(\mathcal{C})$ generated from counterfactual feature selection, the counterfactual example $\x^*$ constructed by the optimal policy of \eqref{eq_4} with $\pi_{\theta}$ defined by \eqref{eq_2} and \eqref{eq_6} is also an optimal solution of the following optimization problem:
\begin{equation}\label{eq_7}
    \max_{\x'}f_{1-y}(\x'),\ s.t.\ \|\x' - \x\|_0 \leq w.
\end{equation}
\end{theorem}
\begin{proof}
Since $\pi_\theta$ is a deterministic policy defined by \eqref{eq_2} and \eqref{eq_6}, Problem \eqref{eq_4} is equivalent to $\max_{\theta=(\p,\q_1,\cdots,\q_s)}r(\theta),\ s.t.\ \|\p\|_0 \leq w$ where $r(\theta) = f_{1-y}(\x')$ and $\x'$ is constructed by $\theta$ whose the $c$th element ($c \in \mathcal{C}$) is defined by $x_c' = \arg\max_v q_{c,v}$ if $p_c = 1$ or $x_c' = x_c$ otherwise. From the definition of $\mathcal{V}(\mathcal{C})$ in counterfactual feature selection, we know that $x_c \not\in \mathcal{V}(c)$ for any $c \in \mathcal{C}$, implying that $\|\p\|_0 = \|\x' - \x\|_0$. Therefore, Problem \eqref{eq_4} is equivalent to $\max_{\x'}f_{1-y}(\x'),\ s.t.\ \|\x' - \x\|_0 \leq w$, so that the optimal policy of \eqref{eq_4} is able to construct an optimal solution of \eqref{eq_7}.
\end{proof}

From Theorem \ref{th_1} we know that the optimization approach as shown in \eqref{eq_7} for finding counterfactual examples can be reformulated as a reinforcement learning problem with the policy defined by \eqref{eq_2} and \eqref{eq_6}. Because optimizing $\eqref{eq_4}$ with $\mathcal{P}$ and $\mathcal{Q}_c$ defined by \eqref{eq_6} is NP hard, we do the relaxation by replacing \eqref{eq_6} with its ``soft'' formulation \eqref{eq_3}. This means that our approach by solving Problem \eqref{eq_5} actually approximately solves Problem \eqref{eq_7}.

\subsection{Counterfactual Example Selection}

Given a set of counterfactual examples generated in the previous step, the goal of counterfactual example selection is to pick a set of diverse counterfactual examples such that the proximity of each counterfactual example is as high as possible (or the distance between each counterfactual example and the original query is as small as possible). ``Diverse'' means we try to select different sets of features for different counterfactual examples. Generating diverse counterfactual examples can help us to get a better understanding of machine learning prediction. This problem can also be formulated as an optimization problem. But due to the concern about computational cost, we propose a simple greedy algorithm for this problem as shown in Algorithm \ref{alg_diverse}. We first sort the examples in $\mathcal{E}$ in descending order based on the sum of categorical proximity and continuous proximity, i.e., 
$$\text{Proximity} = -\|\x_{cat}' - \x_{cat}\|_0 - \|(\x_{con}' - \x_{con}) / \m_{con}\|_1,$$ 
where $\x_{cat}$ and $\x_{con}$ are the categorical and continuous features respectively, and $\m_{con}$ is the median of $\x_{con}$ over the training data. We then try to remove ``duplicate'' examples from the sorted $\mathcal{E}$ in a greedy manner to make the counterfactual feature columns (their feature values are different from $\x$) of the selected examples less overlapped with each other. For each example $\x'$ in the sorted $\mathcal{E}$, let $\mathcal{D}(\x')$ be the set of the feature columns in $\x'$ whose values are different from $\x$. If there exists any feature column in $\mathcal{D}(\x')$ that appears more than $K$ times in the sets $\mathcal{D}(\z)$ for $\z$ comes before $\x'$ in the sorted $\mathcal{E}$, then example $\x'$ is ignored. For practical applications, one can pick top 3 examples in $\mathcal{R}$ returned by Algorithm \ref{alg_diverse}.

\begin{algorithm}
\caption{Counterfactual example selection}
\label{alg_diverse}
\begin{algorithmic}
    \REQUIRE \ A set of counterfactual examples $\mathcal{E}$ obtained by the counterfactual feature optimization step (e.g., Algorithm \ref{alg_feature_optimization}), and parameter $K$.
    \ENSURE ~~
        \STATE $1)$ Sort $\mathcal{E}$ in descending order based on the proximity scores $-\|\x_{cat}' - \x_{cat}\|_0 - \|(\x_{con}' - \x_{con}) / \m_{con}\|_1$ for $\x' \in \mathcal{E}$;
        \STATE $2)$ Initialize counts $c[f]$ for each feature $f$ to be 0 and set $\mathcal{R}$ to be an empty list;
        \FOR {$\x' \in \text{ the sorted } \mathcal{E}$}
            \STATE $3)$ Let $\mathcal{D}(\x')$ be the set of features $f$ such that $\x'[f] \neq \x[f]$;
            \STATE $4)$ For each feature $f \in \mathcal{D}(\x')$, if there is no $c[f] >= K$, then add $\x'$ into $\mathcal{R}$ and update $c[f] = c[f] + 1$ for $f \in \mathcal{D}(\x')$;
        \ENDFOR
        \STATE $5)$ Return $\mathcal{R}$.
\end{algorithmic}
\end{algorithm}

\textbf{Remark.} The RL-based method only considers categorical proximity since continuous features are converted into categorical ones by discretization. Algorithm \ref{alg_diverse} addresses this issue by picking the best examples from those generated by the RL-based method according to the sum of categorical proximity and continuous proximity. 

\subsection{Continuous Feature Fine-tuning}

One concern about discretizing continuous features is that the generated counterfactual examples may have better or more actionable values for those continuous features if they are optimized directly instead of being discretized via ordinal encoding. The continuous feature fine-tuning module can be applied if one wants to optimize continuous proximity further to generate more actionable explanations. We propose a new gradient-less descent (GLD) method based on \cite{golovin2020gradientless} for refining continuous counterfactual features. 

Given a counterfactual example $\x'$ generated by the previous stage, let $\mathcal{C}$ be the set of the continuous features such that $\x'[i] \neq \x[i]$ for each $i \in \mathcal{C}$, i.e., the set of the continuous features that are modified to construct $\x'$. Without loss of generality, we assume that the value of each continuous feature lies in $[0,1]$. The goal is to optimize the values $\x'[i]$ for $i \in \mathcal{C}$ such that the new counterfactual example $\hat{\x}$ satisfies $f_{1-y}(\hat{\x}) > 0.5$ and the difference between $\hat{\x}$ and $\x$ on the continuous features, i.e., $\sum_{i \in \mathcal{C}}|\hat{\x}[i] - \x[i]|$, is as small as possible. In other words, we want to solve the following problem:
\begin{equation}\label{a_eq_1}
\begin{aligned}
    &\min_{\z} \frac{1}{|\mathcal{C}|}\sum_{i \in \mathcal{C}}\left|\z[i] - \x[i]\right|, \\
    &s.t.\ f_{1-y}(\z) > 0.5,\ \z[i] = \x'[i]\ \forall i \not\in \mathcal{C},\ \z[i] \in [0,1]\ \forall i \in \mathcal{C}.
\end{aligned}
\end{equation}
This problem can be approximately solved via GLD-search as shown in Algorithm \ref{a_alg_gld_optimization} that takes a binary sweep across an interval $[r,R]$.
\begin{algorithm}
\caption{The GLD method for continuous feature fine-tuning}
\label{a_alg_gld_optimization}
\begin{algorithmic}
    \REQUIRE \ Classifier $f(\cdot)$, query instance $(\x,y)$, CF example $\x'$, maximum search radius $R$,  minimum search radius $r$, and number of iterations per epoch $T$.
    \ENSURE ~~
        \STATE $1)$ Set $K = \log(R / r)$ and let $\mathcal{X}$ be an empty list;
        \STATE $2)$ Set $\mathcal{C} = \{i : \text{ feature } i \text{ is continuous and } \x'[i] \neq \x[i]\}$ and set $\z = \x'$;
        \FOR {$t = 1,\cdots,T$}
            \STATE $3)$ Let $\mathcal{T}_t$ be an empty list;
            \FOR {$k = 1,\cdots,K$}
                \STATE $4)$ Set $r_k = 2^{-k}R$;
                \STATE $5)$ Set $\z_k[i] = \text{Clip}(\z[i] + a * r_k, 0, 1)$ where $a \sim \mathcal{N}(0,1)$ for each $i \in \mathcal{C}$;
                \STATE $6)$ Add $\z_k$ into $\mathcal{T}_t$ if $f_{1-y}(\z_k) > 0.5$;
            \ENDFOR
            \STATE $7)$ Update 
            $\z$ to be the best solution of \eqref{a_eq_1} in $\mathcal{T}_t$, and add $\z$ into the solution list $\mathcal{X}$;
        \ENDFOR
        \STATE $8)$ Return $\z \in \mathcal{X}$ that minimizes \eqref{a_eq_1}.
\end{algorithmic}
\end{algorithm}

\textbf{Remark.} Based on the same idea, the GLD method can also be applied for counterfactual feature optimization, i.e., optimizing 
\begin{equation*}
    \min_{\x'} \frac{1}{d}(\|\x_{cat}' - \x_{cat}\|_0 + \|(\x_{con}' - \x_{con}) / \m_{con}\|_1),\ s.t.\ f_{1-y}(\x') > 0.5.
\end{equation*}
\textbf{This algorithm can be found in Section 3 in the appendix.} We take it as a baseline in the experiments.

\section{Experiments}\label{sec:experiment}

The experiments include three parts: (1) evaluate our approach for generating one counterfactual example, (2) evaluate our approach for generating diverse counterfactual examples, and (3) test our approach on time series classification tasks. \textbf{More detailed ablation study also can be found in Section 2 in Appendix.}

\subsection{Datasets}

To do the evaluation, we use the following five tabular datasets and one time series classification dataset: 1) ``Adult Income'' \cite{Dua:2019} for predicting whether income exceeds \$50K/yr based on census income data, 2) ``Breast Cancer'' \cite{Dua:2019} for classifying whether an instance is malignant or benign, 3) ``COMPAS'' \cite{Andini2018} for predicting which of the bail applicants will recidivate in the next two years. 4) ``Australian Credit'' \cite{Dua:2019} for predicting if the credit application was approved or rejected to a particular customer. 5) ``Titanic\footnote{https://www.kaggle.com/c/titanic/overview}'' for predicting which passengers survived the Titanic shipwreck. 6) ``Taxonomist'' \cite{Ates2018} for classifying different applications based on the collected 563 time series (563 features) per sample. \textbf{Due to space limitation, more experimental results on other datasets can be found in Section 1 in Appendix.}

Each dataset is divided into 80\% training and 20\% test data for training and evaluating machine learning classifiers. We consider three popularly used classifiers, an XGBoost model and an MLP model for tabular classification, and a random forest model for time series classification.

\subsection{Experimental Setups}

Our MACE-RL approach applies Algorithm \ref{alg_nn_search} for CF feature selection, the RL-based method (Algorithm \ref{alg_feature_optimization}) for CF feature optimization, Algorithm \ref{alg_diverse} for CF example selection and Algorithm \ref{a_alg_gld_optimization} for continuous feature fine-tuning. Our MACE-GLD method replaces the RL-based method with the GLD method for CF feature optimization. For Algorithm \ref{alg_nn_search}, we set $K=30$, $s=10$ and $m=3$. Problem \eqref{eq_5} is solved via REINFORCE. We choose ADAM \cite{kingma2017adam} as the optimizer with learning rate $0.1$, batch size $40$ and $15$ epochs. In order to make the REINFORCE algorithm stable, the reward function used here is $r = f_{1-y}(\x') - b$ where $b$ is the median value of $f_{1-y}(\cdot)$ in one batch for variance reduction. The regularization weights $\lambda_1$ and $\lambda_2$ are set to 2. For Algorithm \ref{alg_feature_optimization}, the parameters $w$ and $B$ are set to $8$ and $80$, respectively. For Algorithm \ref{alg_diverse}, the parameter $K$ is set to 3. For Algorithm \ref{a_alg_gld_optimization}, we set $T=20$, $R=0.25$ and $r=0.0005$. These setups work well for all the experiments.

\subsection{Baselines and Metrics}

We compare our approach with Greedy \cite{ates2020counterfactual}, SHAP-based \cite{rathi2019generating}, DiCE \cite{Mothilal2020} and CERTIFAI \cite{Sharma2020}. To measure the performance, we use the validity, sparsity, proximity and diversity metrics as discussed in \cite{Mothilal2020}. Validity is the fraction of valid counterfactual examples returned by a method, e.g., $f_{1-y}(\x') > 0.5$. Sparsity indicates the number of changes between the query $\x$ and counterfactual example $\x'$. Proximity measures the average distance between the counterfactual example and the query instance. We consider the sum of the categorical proximity and continuous proximity, i.e., $-\|\x_{cat}' - \x_{cat}\|_0 - \|(\x_{con}' - \x_{con}) / \m_{con}\|_1$. Diversity measures the fraction of features that are different between any two pair of counterfactual examples (e.g., generate 3 CF examples per query instance). For sparsity, the lower the better. For validity, proximity and diversity, the higher the better.

\begin{table}[!ht]
\caption{The validity, sparsity and proximity for Greedy, SHAP-based, CERTIFAI, MACE-GLD and MACE-RL with the XGBoost classifier.}\vspace{-0.1in}
\label{table_3}
\center
%\begin{small}
\begin{tabular}{|c|c|c|c|c|c|}
  \hline
  \textbf{Validity} & Adult  & Breast & COMPAS & Credit & Titanic \\ \hline
  Greedy      & 0.99 & 1.00 & 0.96 & 0.94 & 1.00 \\ \hline
  SHAP-based  & 0.99 & 1.00 & 0.97 & 0.95 & 1.00 \\ \hline
  CERTIFAI    & 0.95 & 0.99 & 1.00 & 0.97 & 1.00 \\ \hline
  MACE-GLD    & 1.00 & 1.00 & 1.00 & 1.00 & 1.00 \\ \hline
  MACE-RL     & 1.00 & 1.00 & 1.00 & 1.00 & 1.00 \\ \hline\hline
  
  \textbf{Sparsity} & Adult  & Breast & COMPAS & Credit & Titanic \\ \hline
  Greedy      & 1.68 & 1.99 & 1.38 & 1.39 & 1.30 \\ \hline
  SHAP-based  & 1.89 & 2.07 & 1.60 & 1.43 & 1.37 \\ \hline
  CERTIFAI    & \textbf{1.63} & 2.21 & 1.47 & 2.67 & 1.37 \\ \hline
  MACE-GLD    & 1.69 & 2.07 & 1.43 & 1.38 & 1.34 \\ \hline
  MACE-RL     & \textbf{1.63} & \textbf{1.97} & \textbf{1.37} & \textbf{1.30} & \textbf{1.25} \\ \hline\hline
  
  \textbf{Proximity} & Adult  & Breast & COMPAS & Credit & Titanic \\ \hline
  Greedy      & -1.47 & -1.99 & -2.28 & -94.31 & -1.12 \\ \hline
  SHAP-based  & -1.54 & -2.07 & -2.85 & -269.42 & -1.13 \\ \hline
  CERTIFAI    & \textbf{-1.11} & -2.21 & \textbf{-1.94} & -13.51 & -1.12 \\ \hline
  MACE-GLD    & -1.29 & -2.07 & -1.98 & -1.17 & -0.92 \\ \hline
  MACE-RL     & -1.22 & \textbf{-1.97} & \textbf{-1.94} & \textbf{-1.12} & \textbf{-0.76} \\ \hline
\end{tabular}
%\end{small}
\end{table}

\subsection{Experimental Results}

The first experiment compares the methods mentioned above when generating one counterfactual example per query instance. Table \ref{table_3} demonstrates the validity, sparsity and proximity for Greedy, SHAP-based, CERTIFAI, MACE-GLD and MACE-RL with the XGBoost classifier. DiCE is not included here because it cannot handle the XGBoost classifier. Clearly, we can observe that MACE-GLD and MACE-RL have the highest validity, which can successfully generate counterfactual examples for all test instances. Greedy and SHAP-based can obtain good sparsity metrics but low proximity because they don't consider optimizing proximity during counterfactual example generation. MACE-RL obtains the lowest sparsity for all the datasets. For proximity, CERTIFAI and MACE-GLD have similar performance, while MACE-GLD is better than CERTIFAI in ``Breast'', ``CREDIT'' and ``Titanic'' (we treat all the features in ``Breast'' as categorical ones). MACE-RL achieves the highest proximity for all the datasets except ``Adult''.

Table \ref{table_5} demonstrates the validity, sparsity and proximity for Greedy, SHAP-based, CERTIFAI, DiCE, MACE-GLD and MACE-RL with the MLP classifier. DiCE, MACE-GLD and MACE-RL are able to find valid counterfactual examples for all test instances, outperforming the other approaches. But in terms of sparsity, DiCE has much larger sparsity than our methods, e.g., 2.93 for ``Adult'' and 2.10 for ``COMPAS'', while our MACE-GLD/RL achieves 1.57/1.52 for ``Adult'' and 1.53/1.44 for ``COMPAS''. For ``Titanic'', Greedy, SHAP-based, CERTIFAI, MACE-GLD and MACE-RL have similar performance in terms of sparsity, while CERTIFAI and MACE-RL achieve the best performance in terms of proximity, which is different from the results with the XGBoost classifier. Clearly, MACE-RL obtains the lowest sparsity and highest proximity in almost all the datasets. 

Table \ref{table_6_3} compares the running time per query for all these methods. DiCE is the slowest method which takes about 3 seconds per query, making it impractical for real applications. MACE-RL is about two times faster than MACE-GLD and eight times faster than CERTIFAI, and achieves comparable speed as Greedy and SHAP-based while obtaining better performance in terms of validity, sparsity and proximity. 

\begin{table}[!ht]
\caption{The validity, sparsity and proximity for Greedy, SHAP-based, CERTIFAI, DiCE, MACE-GLD and MACE-RL with the MLP classifier.}\vspace{-0.1in}
\label{table_5}
\center
%\begin{small}
\begin{tabular}{|c|c|c|c|c|c|}
  \hline
  \textbf{Validity} & Adult  & Breast & COMPAS & Credit & Titanic \\ \hline
  Greedy      & 1.00 & 1.00 & 0.98 & 0.98 & 1.00 \\ \hline
  SHAP-based  & 0.99 & 1.00 & 0.98 & 0.98 & 1.00 \\ \hline
  CERTIFAI    & 0.96 & 0.94 & 1.00 & 0.98 & 1.00 \\ \hline
  DiCE        & 1.00 & 1.00 & 1.00 & 1.00 & 1.00 \\ \hline
  MACE-GLD    & 1.00 & 1.00 & 1.00 & 1.00 & 1.00 \\ \hline
  MACE-RL     & 1.00 & 1.00 & 1.00 & 1.00 & 1.00 \\ \hline\hline
  
  \textbf{Sparsity} & Adult  & Breast & COMPAS & Credit & Titanic \\ \hline
  Greedy      & 1.53 & 2.89 & 1.52 & 1.30 & \textbf{1.27} \\ \hline
  SHAP-based  & 1.71 & 2.92 & 1.74 & 1.28 & \textbf{1.27} \\ \hline
  CERTIFAI    & 1.57 & 2.96 & 1.47 & 2.11 & 1.28 \\ \hline
  DiCE        & 2.93 & 5.37 & 2.10 & 11.84 & 3.24 \\ \hline
  MACE-GLD    & 1.57 & 2.91 & 1.53 & 1.27 & 1.28 \\ \hline
  MACE-RL     & \textbf{1.52} & \textbf{2.82} & \textbf{1.44} & \textbf{1.25} & \textbf{1.27} \\ \hline\hline
  
  \textbf{Proximity} & Adult  & Breast & COMPAS & Credit & Titanic \\ \hline
  Greedy      & -1.43 & -2.89 & -2.76 & -1.30 & -1.27 \\ \hline
  SHAP-based  & -1.48 & -2.92 & -2.93 & -1.28 & -1.27 \\ \hline
  CERTIFAI    & -1.43 & -2.96 & -1.68 & -1.92 & \textbf{-1.18} \\ \hline
  DiCE        & -1.87 & -5.37 & -4.93 & -5023.03 & -2.02 \\ \hline
  MACE-GLD    & -1.31 & -2.91 & -1.84 & -1.25 & -1.21 \\ \hline
  MACE-RL     & \textbf{-1.20} & \textbf{-2.82} & \textbf{-1.65} & \textbf{-1.24} & \textbf{-1.18} \\ \hline
\end{tabular}
%\end{small}
\end{table}

\begin{table}[!ht]
\caption{The running time for generating counterfactual examples (seconds per query).}\vspace{-0.1in}
\label{table_6_3}
\center
%\begin{small}
\begin{tabular}{|c|c|c|c|c|c|}
  \hline
   & Adult  & Breast & COMPAS & Credit & Titanic \\ \hline
  Greedy      & 0.030 & 0.056 & 0.024 & 0.058 & 0.032 \\ \hline
  SHAP-based  & 0.106 & 0.226 & 0.041 & 0.645 & 0.064 \\ \hline
  CERTIFAI    & 0.786 & 0.703 & 0.957 & 0.912 & 1.145 \\ \hline
  DiCE        & 2.815 & 5.472 & 3.991 & 2.928 & 2.070 \\ \hline
  MACE-GLD    & 0.254 & 0.236 & 0.234 & 0.314 & 0.262 \\ \hline
  MACE-RL     & 0.129 & 0.110 & 0.123 & 0.161 & 0.131 \\ \hline
\end{tabular}
%\end{small}
\end{table}

\begin{figure}[!ht]
%\vspace{-0.1in}
    \centering
    \begin{subfigure}[b]{0.49\linewidth}
        \includegraphics[width=\linewidth]{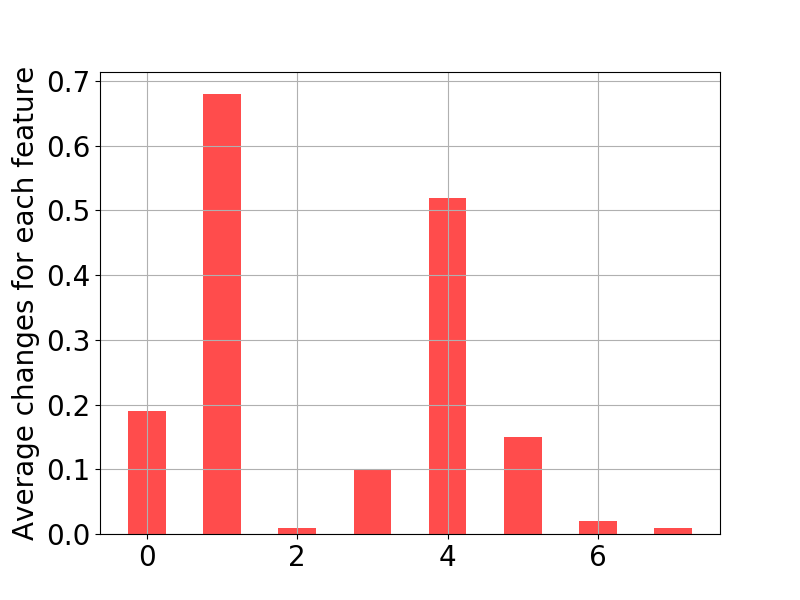}
        \caption{Average number of changes for each feature}
        \label{figure_2}
    \end{subfigure}
    \begin{subfigure}[b]{0.49\linewidth}
        \includegraphics[width=\linewidth]{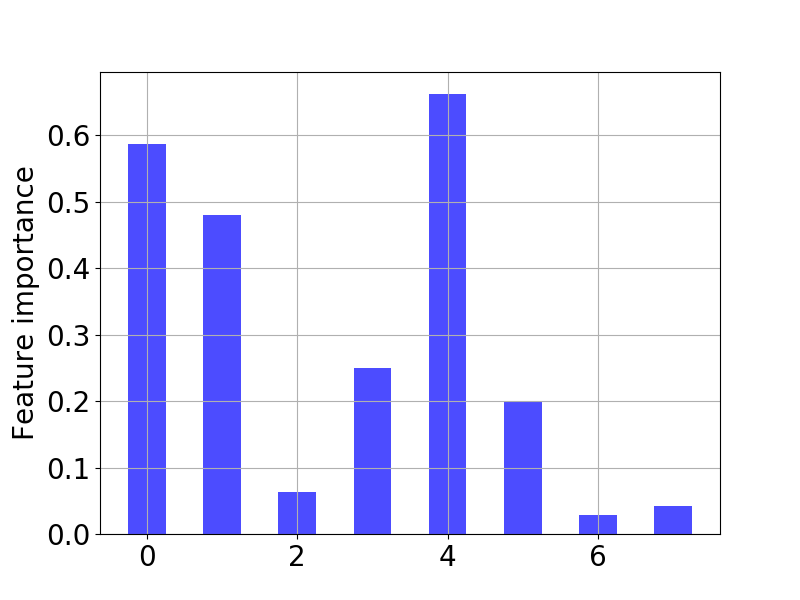}
        \caption{Feature importance scores based on SHAP}
        \label{figure_3}
    \end{subfigure}
    %\vspace{-0.1in}
    \caption{Feature importance comparison. Features 0 to 7 are ``Age'', ``Education'', ``Gender'', ``Working hours'', ``Marital'', ``Occupation'', ``Race'' and ``Workclass'', respectively.}
    \label{fig:feature_importance}
    %\vspace{-0.1in}
\end{figure}

We also compute the average number of the changes for each feature to construct counterfactual example $\x'$ and compare it with the feature importance of the query $\x$. The feature importance is computed by taking the average of the SHAP values of test samples. Figure \ref{fig:feature_importance} shows this comparison with the Adult dataset where features 0 to 7 are ``Age'', ``Education'', ``Gender'', ``Working hours'', ``Marital'', ``Occupation'', ``Race'' and ``Workclass'', respectively.
Clearly, the average feature changes are consistent with the feature importance, e.g., ``Age'', ``Education'' and ``Marital'' are more important, and ``Gender'', ``Race'' and ``Workclass'' are less important.

\begin{table*}[!ht]
\caption{Diverse counterfactual examples generated by our method on the Adult dataset.}\vspace{-0.1in}
\label{table_9}
\center
%\begin{small}
\begin{tabular}{|c|c|c|c|c|c|c|c|c|c|}
  \hline
   & Label & Age & Workclass & Education & Marital & Occupation & Race & Gender & Working hours \\ \hline\hline
  Original & $>50K$      & 28 & Private & Bachelors & Married & Professional & Other & Female & 40 \\ \hline
  Counterfactual 1 & $<=50K$  & 28 & Private & Bachelors & \textbf{Single} & Professional & Other & Female & 40 \\ \hline
  Counterfactual 2 & $<=50K$  & 28 & Private & \textbf{School} & Married & Professional & Other & Female & 40 \\ \hline
  Counterfactual 3 & $<=50K$  & \textbf{23} & Private & Bachelors & Married & Professional & Other & Female & 40 \\ \hline 
  Top 4 actions & - & 23 & - & School & Single & Sales & - & - & -  \\ \hline\hline

  Original & $<=50K$      & 40 & Private & Assoc & Married & Blue-Collar & Other & Male & 40 \\ \hline
  Counterfactual 1 & $>50K$  & 40 & Private & Assoc & Married & \textbf{Professional
} & Other & Male & 40 \\ \hline
  Counterfactual 2 & $>50K$  & 40 & Private & Assoc & Married & Blue-Collar & Other & Male & \textbf{50} \\ \hline
  Counterfactual 3 & $>50K$  & 40 & Private & \textbf{Doctorate} & Married & \textbf{White-Collar} & Other & Male & 40 \\ \hline
  Top 4 actions & - & - & - & Doctorate & - & White-Collar & White & - & 50  \\ \hline
\end{tabular}
%\end{small}
\end{table*}

\begin{table*}[!ht]
\caption{Diverse counterfactual examples generated by our method on the Titanic dataset.}\vspace{-0.1in}
\label{table_10}
\center
%\begin{small}
\begin{tabular}{|c|c|c|c|c|c|c|c|c|}
  \hline
   & Label & Ticket class & Sex & Age & SibSp & Parch & Fare & Embarked \\ \hline\hline
  Original & survival      & 1 & female & 35 & 1 & 0 & 3.9722 & S \\ \hline
  Counterfactual 1 & not survival  & \textbf{3} & \textbf{male} & 35 & 1 & 0 & 3.9722 & S \\ \hline
  Counterfactual 2 & not survival  & \textbf{3} & female & 35 & 1 & 0 & \textbf{3.2959} & S \\ \hline
  Counterfactual 3 & not survival  & 1 & \textbf{male} & \textbf{46} & 1 & 0 & 3.9722 & S \\ \hline
  Top 4 actions & - & 3 & male & 46 & - & - & 3.2959 & - \\ \hline\hline

  Original & not survival      & 2 & male & 35 & 0 & 0 & 3.2581 & S \\ \hline
  Counterfactual 1 & survival  & 2 & \textbf{female} & 35 & 0 & 0 & 3.2581 & S \\ \hline
  Counterfactual 2 & survival  & \textbf{1} & male & 35 & 0 & 0 & 3.2581 & S \\ \hline
  Counterfactual 3 & survival  & 2 & \textbf{female} & \textbf{32} & 0 & 0 & 3.2581 & S \\ \hline
  Top 4 actions & - & 1 & female & 32 & 1 & - & - & - \\ \hline
\end{tabular}
%\end{small}
\end{table*}

\begin{table}[!ht]
\caption{The sparsity, diversity and running time for the generated diverse CF examples with the MLP classifier.}%\vspace{-0.1in}
\label{table_11}
\center
\begin{small}
\begin{tabular}{|c|c|c|c|c|c|}
  \hline
  \textbf{Sparsity} & Adult  & Breast & COMPAS & Credit & Titanic \\ \hline
  CERTIFAI    & 1.86 & 3.33 & 1.94 & 2.42 & 1.75 \\ \hline
  DiCE        & 3.14 & 5.81 & 2.27 & 11.99 & 3.69 \\ \hline
  MACE-GLD    & 1.75 & \textbf{3.02} & 1.68 & 1.74 & 1.54 \\ \hline
  MACE-RL     & \textbf{1.71} & 3.08 & \textbf{1.58} & \textbf{1.73} & \textbf{1.48} \\ \hline\hline
  
  \textbf{Diversity} & Adult  & Breast & COMPAS & Credit & Titanic \\ \hline
  CERTIFAI    & 0.54 & 0.33 & 0.44 & 0.41 & 0.49 \\ \hline
  DiCE        & \textbf{0.96} & \textbf{0.98} & \textbf{0.88} & \textbf{0.67} & \textbf{0.86} \\ \hline
  MACE-GLD    & 0.70 & 0.56 & 0.69 & 0.51 & 0.65 \\ \hline
  MACE-RL     & 0.79 & 0.72 & 0.80 & 0.54 & 0.71 \\ \hline\hline
  
  \textbf{Running Time (s)} & Adult  & Breast & COMPAS & Credit & Titanic \\ \hline
  CERTIFAI    & 0.770 & 0.756 & 0.969 & 0.933 & 1.147 \\ \hline
  DiCE        & 7.096 & 8.240 & 5.815 & 5.039 & 3.707 \\ \hline
  MACE-GLD    & 0.272 & 0.246 & 0.245 & 0.319 & 0.268 \\ \hline
  MACE-RL     & \textbf{0.133} & \textbf{0.112} & \textbf{0.126} & \textbf{0.180} & \textbf{0.141} \\ \hline
\end{tabular}
\end{small}
\vspace{-0.1in}
\end{table}

The second experiment compares CERTIFAI, DiCE, MACE-GLD and MACE-RL for generating diverse counterfactual examples. Greedy and SHAP-based are not compared because they cannot generate diverse CF examples. Table \ref{table_11} shows the sparsity, diversity and running time. Our MACE-GLD/RL methods obtain the lowest sparsity while DiCE achieves the highest diversity. DiCE is much slower than CERTIFAI, MACE-GLD and MACE-RL. MACE-RL makes a good trade-off between sparsity and diversity with much less running time per query.

Tables \ref{table_9} and \ref{table_10} show some diverse counterfactual examples generated by MACE-RL. For simplicity, we didn't fine-tune the continuous features here. For the Adult dataset, the counterfactual examples show that studying for an advanced degree or spending more time at work can lead to a higher income. It also shows a less obvious counterfactual example such as getting married for a higher income, which is also mentioned in the paper \cite{Mothilal2020}. This kind of counterfactual example is generated due to correlations in the dataset that married people having a higher income. For the Titanic dataset, the results show that the persons who are female or have higher ticket classes have a bigger chance to survive from the disaster.

Besides the counterfactual examples, our approach can also take counterfactual features extracted from the policy learned by Algorithm \ref{alg_feature_optimization} as the explanation to explore ``what-if'' scenarios. In Tables \ref{table_9} and \ref{table_10}, the top-4 selected feature columns and top-1 feature values of the optimal policy are listed. For the second example in Table \ref{table_9}, we got these counterfactual features ``Education->Doctorate, Working hours->50, Occupation->White-Collar'', meaning that he/she will have a higher income if he/she has a higher degree, longer working hours and a White-Collar occupation. For the first example in Table \ref{table_10}, we got ``Sex->male, Ticket class->3, Age->46'', which means gender being female or having a higher class leads to a bigger survival chance. These counterfactual explanations help detect bias in the dataset which may reflect some bias in our society under certain scenarios. For example, women were more likely to survive because society's values dictate that women and children should be saved first. And rich passengers were more likely to survive because they had rooms on the higher classes of the ship so that they may have the privilege to get to the boats more easily. 

\begin{table}[!ht]
\caption{The sparsity and running time (seconds per query) for generating CF examples with time series data. The original label is ``ft''. The validity metrics are 1.0 for all the setups.}\vspace{-0.1in}
\label{table_14}
\center
\begin{small}
\begin{tabular}{|c|c|c|c|c|c|c|c|c|}
  \hline
  Desired label & \multicolumn{2}{|c|}{bt} & \multicolumn{2}{|c|}{cg} & \multicolumn{2}{|c|}{kripke}  \\ \hline
  & Sparsity & Time & Sparsity & Time & Sparsity & Time  \\ \hline
  Greedy    & 12.2 & 19.6 & 6.6 & 9.9 & 15.8 & 25.9  \\ \hline
  MACE-GLD  & 16.3 & 1.8 & 8.0 & 1.8 & 21.1 & 1.8  \\ \hline
  MACE-RL   & 14.5 & 1.8 & 7.1 & 1.7 & 18.8 & 1.8  \\ \hline\hline
  
  Desired label & \multicolumn{2}{|c|}{lu} & \multicolumn{2}{|c|}{mg} & \multicolumn{2}{|c|}{miniAMR} \\ \hline
  & Sparsity & Time & Sparsity & Time & Sparsity & Time  \\ \hline
  Greedy    & 13.3 & 22.1 & 7.4 & 10.9 & 19.6 & 31.7 \\ \hline
  MACE-GLD  & 13.1 & 1.8 & 9.4 & 1.8 & 19.6 & 1.8 \\ \hline
  MACE-RL   & 14.0 & 1.8 & 8.3 & 1.7 & 19.3 & 1.9 \\ \hline
\end{tabular}
\end{small}
%\vspace{-0.1in}
\end{table}

The third experiment aims to demonstrate that the proposed MACE-GLD and MACE-RL methods can also be applied to time series classification problems. This experiment was conducted on the Taxonomist dataset (each sample contains 563 time series), where the query label is ``ft'' and we take the other labels as the desired labels and generate counterfactual examples for all the desired labels. The counterfactual examples are constructed by substituting several time series (features) as discussed in Section 2.1 in Appendix. We only compare our methods with Greedy since the other methods cannot handle time series data directly. Table \ref{table_14} shows the sparsity and the running time for these methods, from which we observe that our methods are more than 10 times faster than Greedy while achieving similar sparsity metrics for different setups. This result also shows that our methods can handle high dimensional data efficiently, making our methods able to deploy in real applications. 

\section{Conclusions}

Most the previous methods formulate counterfactual explanation as a continuous optimization problem by assuming the underlying ML model is differentiable, converting categorical features into continuous ones and imposing additional constraints on features to guarantee real-world plausibility. Those approaches lead to promising results but still have a gap for practical applications, e.g., the widely-used XGBoost is non-differentiable, and treating categorical features as continuous ones complicates the optimization and leads to high computational cost. 
We addressed these issues by proposing an efficient framework via a newly designed pipeline that consists of counterfactual feature selection, counterfactual feature optimization, counterfactual example selection and continuous feature fine-tuning. Our approach is model-agnostic and able to deal with all kinds of machine learning classifiers. Our extensive experimental results on public datasets validated the effectiveness of our approach. Finally, based on our MACE framework, we also built a practical counterfactual explanation toolkit (see more details in Appendix) for our internal use cases and applications. 

\newpage
\bibliographystyle{ACM-Reference-Format}
\bibliography{ref}

\end{document}